\documentclass{article}
\usepackage{ijcai22}

\usepackage{times}
\usepackage{soul}
\usepackage{url}
\usepackage[hidelinks]{hyperref}
\usepackage[utf8]{inputenc}
\usepackage[small]{caption}
\usepackage{graphicx}
\usepackage{amsmath}
\usepackage{amsthm}
\usepackage{booktabs}
\usepackage{algorithm}
\usepackage{enumitem}
\usepackage{algorithmic}
\usepackage{multirow}
\usepackage[symbol]{footmisc}
\title{Communicative Subgraph Representation Learning for Multi-Relational Inductive Drug-Gene Interaction Prediction}

\author{
Jiahua Rao$^{1,\dagger}$
\and
Shuangjia Zheng$^{1,4,*}$\and
Sijie Mai$^2$\And
Yuedong Yang$^{1,3,*}$
\affiliations
$^1$School of Computer Science and Engineering, Sun Yat-sen University\\
$^2$School of Electronic and Information Technology, Sun Yat-sen University\\
$^3$Key Laboratory of Machine Intelligence and Advanced Computing, Sun Yat-sen University\\
$^4$Galixir Technologies Ltd, Beijing
\emails
\{raojh6,zhengshj9,maisj\}@mail2.sysu.edu.cn,
yangyd25@mail.sysu.edu.cn
}

\begin{document}

\maketitle
\renewcommand{\thefootnote}{\fnsymbol{footnote}}
\footnotetext[1]{Corresponding authors.}
\footnotetext[2]{Work done during an internship at Galixir.}

\begin{abstract}
    Illuminating the interconnections between drugs and genes is an important topic in drug development and precision medicine.  Currently, computational predictions of drug-gene interactions mainly focus on the binding interactions without considering other relation types like agonist, antagonist, etc. In addition,  existing methods either heavily rely on high-quality domain features or are intrinsically transductive, which limits the capacity of models to generalize to drugs/genes that lack external information or are unseen during the training process. To address these problems, we propose a novel \textbf{Co}mmunicative \textbf{S}ubgraph representation learning for \textbf{M}ulti-relational \textbf{I}nductive drug-\textbf{G}ene interactions prediction (CoSMIG), where the predictions of drug-gene relations are made through subgraph patterns, and thus are naturally inductive for unseen drugs/genes without retraining or utilizing external domain features. Moreover, the model strengthened the relations on the drug-gene graph through a communicative message passing mechanism. To evaluate our method, we compiled two new benchmark datasets from DrugBank and DGIdb. The comprehensive experiments on the two datasets showed that our method outperformed state-of-the-art baselines in the transductive scenarios and achieved superior performance in the inductive ones. Further experimental analysis including LINCS experimental validation and literature verification also demonstrated the value of our model. 
    
    %we validate that the predicted increase/decrease of gene expression caused by drugs are highly consistent with the experimental values on the LINCS database, and demonstrate the new potential drug-gene interactions that are not in the current datasets but have been reported in the literature. 
\end{abstract}

\section{Introduction}

Illuminating the interactions between drugs and genes is crucial in the drug discovery or repositioning process~\cite{pritchard2017enhancing}, aiming to discover effective therapeutic drugs or identify new druggable targets~\cite{strittmatter2014overcoming} and “off-targets” that may cause adverse drug reactions~\cite{malki2020drug}. Although experimental studies remain the most reliable approach for determining drug-gene interactions (DGIs), the characterization of all possible DGIs is a daunting task due to the vast costs involved in experiments~\cite{stachel2014maximizing}. Hence, computational prediction has emerged as an attractive alternative to discover new drug-gene interactions.

Currently, computational predictions of DGIs mainly focus on the binding interaction between drugs and genes to discover drugs with a high binding affinity toward their target \cite{tsubaki2019compound,zheng2020predicting,bagherian2021machine}. However, these methods are unable to illuminate the multiple relational information (i.e., agonist, antagonist, etc.), which is also important for revealing the mechanism behind the interplay between drugs and genes \cite{cotto2018dgidb}. For example, an agonist is a drug that activates the receptor to produce a biological response, while an antagonist is a drug that blocks agonist-mediated responses upon binding to a target gene. Therefore, it is critical to determine the multi-relational interactions between drugs and genes.

In addition, the existing methods fell roughly into two categories: feature-based and network-based methods. The feature-based methods \cite{ozturk2018deepdta,torng2019graph} mainly build the embedding function (i.e. graph representation) with the manually crafted features but often lead to inferior performance when high-quality domain features are not available. Furthermore, although much effort has been devoted to developing the network-based methods \cite{luo2017network,zeng2019deepdr} without domain features, these methods are intrinsically transductive, where the learned latent features cannot generalize to drugs/genes unseen during the training process. To make inductive predictions for unseen entities, one common scenario in the recommendation system studies, scientists utilize subgraphs around the entity pairs without leveraging any global information specific to the whole graph \cite{Zhang2020Inductive,teru2020inductive}. Such local subgraphs contain rich graph pattern information, enabling accurate predictions of target interactions on both the transductive and inductive scenarios.

In this study, we propose a novel Communicative Subgraph representation learning for Multi-relational Inductive drug-Gene interactions prediction (CoSMIG) to address the above problems. We firstly compiled two new challenging benchmarks for multi-relational DGI prediction. To the best of our knowledge, this is the first benchmark of this kind. Furthermore, our model is naturally inductive because we extracted a subgraph by the random walk with restart algorithm for each training interaction without leveraging any domain and global information specific to the drug-gene graph. Therefore, our model could be applied to unseen drugs/genes without retraining. And we also proposed a communicative message passing mechanism to strengthen the role of multi-relational information on the drug-gene graph for accurate multi-relational DGI predictions. Finally, the proposed model is shown to outperform state-of-the-art approaches over two public DGI datasets on both transductive and inductive settings. An independent test on the LINCS L1000 dataset indicated that our predicted increases/decreases of gene expression caused by drugs are highly consistent with the experimental values. More importantly, our method was proven to be able to identify new drug-gene interactions that are not in the current datasets but have been reported in the literature, demonstrating that our model may provide new insights into the understanding of the mechanisms of DGIs.

In brief, the main contributions are listed below:
\begin{itemize}[leftmargin=*]
    \item We proposed a novel inductive subgraph representation learning framework without leveraging any domain and global information of the given drug-gene graph, which could be applied to unseen drugs/genes without retraining.
    \item We made accurate multi-relational DGI predictions by strengthening multi-relational information through our communicative message passing mechanism.
    \item We compiled two new challenging benchmarks for multi-relational DGI prediction and conducted extensive experiments to demonstrate the effectiveness of CoSMIG and its interpretability in understanding the multiple relational information.
\end{itemize}

\section{Related Work}

In this section, we introduced the related works in two areas. First, we focus on reviewing the computational methods for drug-gene binary interaction prediction. Second, we summarized several state-of-the-art recommendation methods that have been adopted as baselines for multiple relational DGIs prediction in our work.

\subsection{Drug-gene interactions}

The key issue with drug-gene interactions is to learn a low-dimensional representation of drugs and genes. Once learned, representations can then be utilized to predict the probability scores of drug-gene binary interactions. Generally, these existing methods can be split into the following two categories: feature-based and network-based methods.

The key idea of feature-based models is to adopt a popular deep learning architecture to represent drugs and proteins/targets. For example, DeepDTA \cite{ozturk2018deepdta} employed a convolution neural network (CNN) to learn representations from the raw protein sequences and SMILES strings and combine them to predict the binding affinities of drug-target interactions, while Graph-CNN \cite{torng2019graph} extracts features from pocket graphs and 2D ligand graphs through the graph convolutional framework. As their input was based on the manually crafted features, they often lead to inferior performance when high-quality domain features are not available.

Another research line is to formulate DGI prediction as a link prediction task within the drug-gene graph \cite{luo2017network,zeng2019deepdr}. They often perform the network diffusion algorithms (i.e. random walk) to obtain informative and low-dimensional vector representations of drugs and genes in the network. After that, they infer new interactions of drugs and genes based on the known drug-gene pairs. However, they are transductive because the trained model could not be generalized to the unseen drugs/genes during the training process.

However, these methods still focused on discovering the binary interaction of drugs and proteins/targets, such as the high binding affinity of drugs and their targets, and fail to determine the interaction types between drugs and genes. In fact, an increasing number of people discovered that the interaction type between drugs and human genes has become an important subject in the research on drug efficacy and human genomics \cite{cotto2018dgidb}. Therefore, illuminating the interconnections between drugs and genes and determining their relationship types could provide new insights into drug repositioning.

\subsection{Recommendation methods}

Taking inspiration from recent progress in recommendation systems, we could re-formulate DGI prediction as a multi-relational recommendation task with drugs and genes respectively representing users and items. 

Matrix completion is one of the widely-used recommender methods, where a matrix with rows and columns respectively representing users and items is used to predict users’ interest in items for filling the missing entries of the rating matrix. However, it is transductive so it often requires retraining when adding a new drug/gene unseen before. To alleviate the inductive problem, Inductive Matrix Completion (IMC)\cite{jain2013provable} has been proposed, which leverages the domain features of users and items. But IMC method still has strong constraints on the quality of the domain features.

Due to the superior graph representation learning abilities, the recommendation tasks have been studied using Graph neural networks (GNNs), showing their superiority and effectiveness \cite{berg2017graph}. For example, \cite{berg2017graph} proposed the graph convolutional matrix completion (GC-MC) which directly applies a GNN to the user-item bipartite graph for extracting user and item latent features. But these kinds of model are still transductive.
 
 A recent inductive graph-based recommender system, PinSage \cite{ying2018graph}, uses content as initial node features and has successfully recommended related pins on Pinterest. Furthermore, IGMC~\cite{Zhang2020Inductive} proposed a graph-level GNN to learn representations for subgraphs and use the subgraph embedding to predict the ratings between users and items. However, these GNN models are limited by the node-to-node message passing mechanism, which cannot explicitly consider the relation embeddings in the embedding propagation layer. This set of methods, together with traditional MF-based methods constitute our baselines.

% \subsubsection{References}

% The references section is headed ``References'', printed in the same
% style as a section heading but without a number. A sample list of
% references is given at the end of these instructions. Use a consistent
% format for references. The reference list should not include publicly unavailable work.

% \subsection{Citations}

% Citations within the text should include the author's last name and
% the year of publication, for example~\cite{gottlob:nonmon}.  Append
% lowercase letters to the year in cases of ambiguity.  Treat multiple
% authors as in the following examples:~\cite{abelson-et-al:scheme}
% or~\cite{bgf:Lixto} (for more than two authors) and
% \cite{brachman-schmolze:kl-one} (for two authors).  If the author
% portion of a citation is obvious, omit it, e.g.,
% Nebel~\shortcite{nebel:jair-2000}.  Collapse multiple citations as
% follows:~\cite{gls:hypertrees,levesque:functional-foundations}.
% \nocite{abelson-et-al:scheme}
% \nocite{bgf:Lixto}
% \nocite{brachman-schmolze:kl-one}
% \nocite{gottlob:nonmon}
% \nocite{gls:hypertrees}
% \nocite{levesque:functional-foundations}
% \nocite{levesque:belief}
% \nocite{nebel:jair-2000}

% \subsection{Footnotes}

% Place footnotes at the bottom of the page in a 9-point font.  Refer to
% them with superscript numbers.\footnote{This is how your footnotes
% should appear.} Separate them from the text by a short
% line.\footnote{Note the line separating these footnotes from the
% text.} Avoid footnotes as much as possible; they interrupt the flow of
% the text.

\begin{figure*}
       \centering
	   \includegraphics[width=1.0\textwidth]{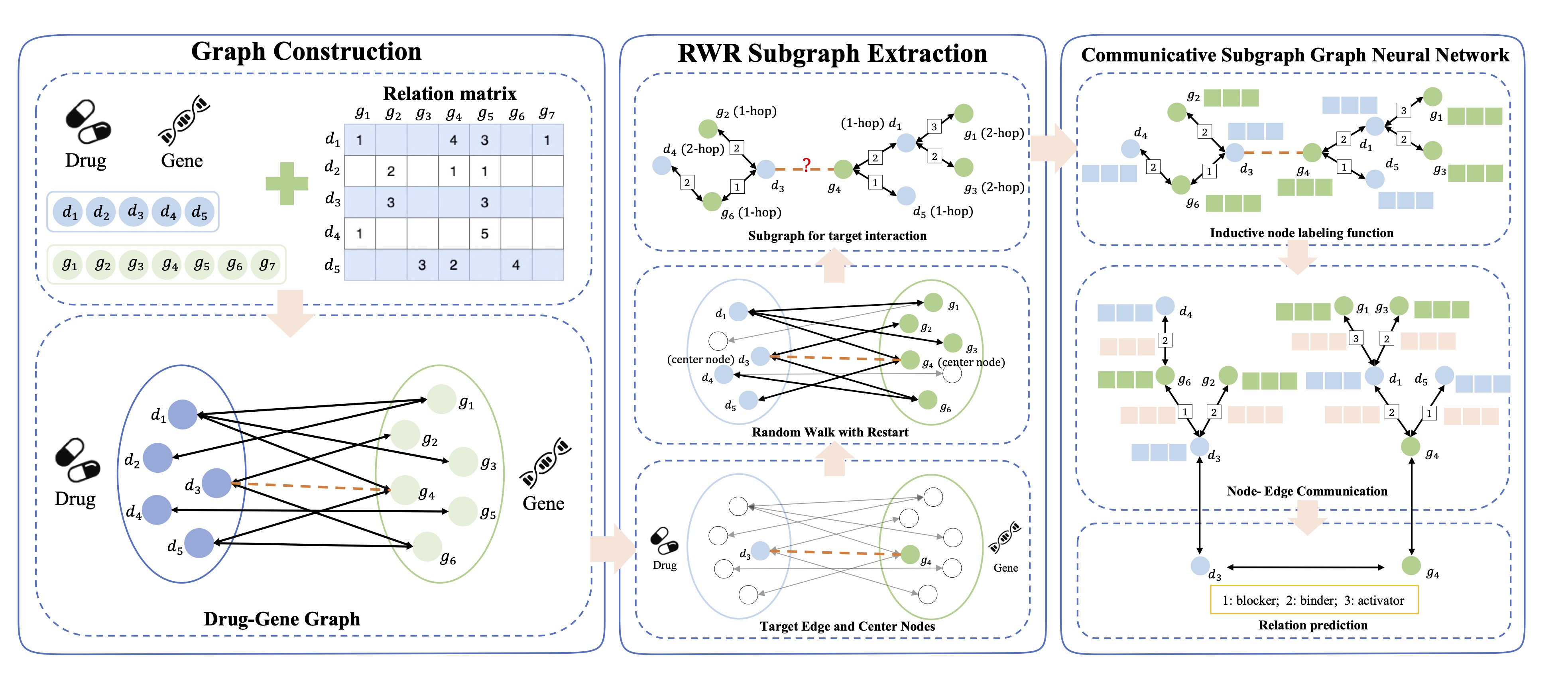}
	   \caption{The framework of CoSMIG. We first extract a subgraph around each interaction and train a communicative subgraph graph neural network to map subgraphs to interactions. Each subgraph is induced by the drug and gene associated with the target interaction as well as their h-hop neighbors (here $h$ = 3). Finally, the learned subgraph embedding of each interaction is used to predict the various interactions between drug and gene.}
\end{figure*}

\section{Methods}

%For our model, the drug-gene interaction data is represented as a graph \{$(d, g, r)$ $|$ $d \in D$, $g \in G$, $r \in R$ \} where $D$, $G$ , and $R$ denote the drug, gene, and relation sets, respectively, and the edge $r$ indicates  an observed interaction between drug $d$ and gene $g$. 
As shown in Figure 1, we predict DGIs through CoSMIG  consisting of three main components: 1) subgraph extraction layer, which extracts the h-hop subgraph by random walk with restart (RWR); 2) embedding layer, which encodes the h-hop subgraph through a graph-level GNN; and 3) prediction layer, which predicts the interactions between two subgraph. 

\subsection{RWR subgraph extraction}

For one given pair of gene $g$ and drug $d$, we first constructed a subgraph by extracting h-hop neighbors around the pair $(d,g)$. Since our model represents DGI through a bipartite graph, h has to be an odd number for drug-gene interaction and h is usually greater than 2 for accurate prediction. Therefore, full extraction of all h-hop neighbors may cause intractably large subgraphs, and we extract subgraphs through the random walk with restart (RWR) instead of the breadth-first search by previous works \cite{Zhang2020Inductive}. 
More specifically, given an interaction between drug and gene, we conduct random walk with restart from the two nodes around the interaction and have to find a solution to the equation:

\begin{equation}
  p = cAD^{-1}p + (1-c)e
\end{equation}

where $c$ is a number in the range $(0,1)$ called restart probability and $p$ is a column vector with $p_i$ denoting the probability at node $i$.  $D$ is the degree matrix of adjacency matrix $A$ with each diagonal value $D_{ii} = \sum_{j}A_{ij}$ . The restart probability controls whether the next walk is jumping to a randomly selected neighbor (with probability $c$) or going to the starting node (with probability $1-c$). For the starting vector e, we set $e_i=1$ if node $i$ is the starting node else 0, and thus the starting vector $e$ allows us to preserve the node’s local topological structure and $AD^{-1}$ allows us to further visit their neighborhoods. 

\subsection{Communicative subgraph Neural Network}

After extracting h-hop subgraph around the pair $(d,g)$, we designed a communicative subgraph neural network, inspired by previous works\cite{song2020communicative,mai2021communicative}, to model the inductive subgraph by iteratively communicating the relation and node embeddings and strengthening the role of multi-relational information. %As our subgraph is a bipartite graph, we define the drug-gene interactions undirected.

Given an extracted subgraph $\mathcal{G}$ with $n$ nodes and $e$ edges, we represent $N$ for the node embeddings, $E$ for the edge embeddings, and $R$ for the relation embeddings. We also defined the adjacency matrix in the subgraph as $A^{ne}$ and $A^{re}$ which represents the node-to-edge and relation type-to-edge adjacency matrix, respectively.

Before feeding to CoSMIG, we first apply an inductive node labeling function to it, which uses different labels to mark nodes’ different roles in the subgraph without leveraging any external domain features and global information. Our node labeling function is defined as $(2i+j)$ where $i$ is the hop number of the node and $j$ is the node type with 0 representing the drug nodes and 1 representing the gene nodes. The one-hot encoding of these node labels will be treated as the initial node features of the subgraph, denoted as $N_0$. The initial relational feature $R_0$ is the one-hot encoding of interaction types and the initial edge features $E_0$ is the initial relation feature of each edge.

Therefore, we firstly mapped the node and edge representations to the same dimensionality $f$:

\begin{equation}
  N^{(0)} = \sigma(N_0 W_{n}^{(0)}), E^{(0)} = \sigma(E_0 W_{e}^{(0)})
\end{equation}

where $\sigma$ denotes the nonlinear activation function, $W_n^{(0)}$ and $W_e^{(0)}$ are the learnable parametric matrix, $N^{(0)} \in R^{n \times f}$ and $E^{(0)} \in R^{e \times f}$ are the transformed node and relation embeddings , respectively.

\textbf{Node embedding aggregation:} In our node-edge interaction mechanism, relation embedding is required for updating the node embedding at each iteration. Firstly, we use the node embedding ($N_i, N_j$) and the relation embedding ($E_{i,j}$) to calculate the weight for the associations $i,j$ through edge attention scores by:

\begin{equation}
    \alpha_{i,j}^{(k)} = \sigma(\sigma([N_i^{(k)} || N_j^{(k)} || E_{i,j}^{(k)}] W_{a_0}^{(k-1)})W_{a_1}^{(k)})
\end{equation}

where $W_{a_0}^{(k)}$ and $W_{a_1}^{(k)}$ denotes the learnable parametric matrix at the iteration $k$ and the operator $||$ denotes feature concatenation .

Then we use the attentive edge embedding $E_{i,j}^{(k-1)}$ to update the node representation:

\begin{equation}
    E_{i,j}^{(k-1)} = \alpha_{i,j}^{(k-1)} E_{i,j}^{(k-1)}
\end{equation}

\begin{equation}
    N_{agg}^{(k)} = A^{ne} E^{(k-1)}
\end{equation}

\begin{equation}
    N^{(k)} = \sigma((N_{agg}^{(k)}+N^{(k-1)})W_n^{(k)})
\end{equation}

where $N_{agg}^{(k)}$ denotes the node aggregation information from its neighbors at the iteration $k$, $W_n^{(k)}$ represents the parametric matrix for node embedding at the iteration $k$, $E_{i,j}^{(k-1)}$ is the attentive edge embedding for edge $(i,j)$. Therefore, By using the relation embedding to update node embedding, the node embedding can aggregate all the relations along with their relative information in the subgraph, providing powerful relational inference ability.

\textbf{Relation embedding updating:} The relation embedding is updated for totally $l-1$ iterations and the node embedding is required to update the relation embedding. We first aggregate information from node to edge and relation to edge by:

\begin{equation}
    E^{(k)}_{agg} = (A^{ne})^{T}N^{(k)} + (A^{re})^{T}R^{(k)}
\end{equation}
where $T$ denotes matrix transpose, $(A^{he})^TN^{(k)}$ aggregates the head information to edge, and $ (A^{re})^{T}R^{(k)}$ aggregates the relation information to edge.

Then we use the aggregation information to update edge representation:

\begin{equation}
    {E^{(k)}}^{'} = \sigma(E^{(k-1)} + \sigma(E_{agg}^{(k)}))
\end{equation}

\begin{equation}
    E^{(k)} = \sigma({E^{(k)}}^{'}W_e^{(k)}+E^{(0)})
\end{equation}

where $\sigma$ denotes nonlinear activation function, $W_e^k$ denotes the learnable parametric matrix, and $E^0$ is the original transformed relation embedding  to perform residual learning. 

\subsection{Relation prediction based on subgraph embedding}

In this section, we pool the node representations into a graph-level feature vector to represent the interaction $(d,g)$. There are many choices such as summing, averaging, SortPooling~\cite{zhang2018end}, DiffPooling \cite{ying2018hierarchical}. In this work, we only concatenate the final representations of the target drug $d$ and gene $g$ as the graph representation:

\begin{equation}
    h = concat(N_d, N_g)
\end{equation}

where $N_d$ and $N_g$  denote the final representations of the target drug and  gene, respectively.  We empirically verified its superior performance because highlighting the target drug and gene is important for DGIs predictions.

After getting the final graph representation, we use an MLP to output the predicted interactions:

\begin{equation}
    \widehat{r} = w^{T}_{y} \sigma(W_hh)
\end{equation}
where $\sigma$ is an activation function, and $w^{T}_{r}$ and $W_h$ are parameters of the MLP for mapping the graph representation $h$ to a scalar prediction $\widehat{r}$.

\subsection{Model training}

Following \cite{Zhang2020Inductive}, we minimize the mean squared error (MSE) between the predictions and the ground truth interactions:

\begin{equation}
    \mathcal{L} = \left( \frac{1}{|{(d,g)|\Omega_{d,g}=1}|} \right) \sum_{(d,g):\Omega_{d,g}=1} {(r_{d,g} - \widehat{r}_{d,g})^2}
\end{equation}
where $r_{d,g}$ and $\widehat{r}_{d,g}$ denote the true interaction and predicted interaction of $(u, v)$, respectively, and $\Omega$ is a 0/1 mask matrix indicating the observed entries of the interaction matrix $R$.

\begin{table}
\centering
\begin{tabular}{lll}
\hline
Dataset  & DrugBank & DGIdb \\
\hline
Number of Drug       & 425  & 1185     \\
Number of Gene       & 11284  & 1664      \\
Interactions    & 80924  & 11366     \\
Interaction types   & 2  & 14     \\
\hline
\end{tabular}
\caption{Statistics of two Drug-Gene Interaction datasets}
\label{tab:plain}
\vspace{-1.0em}
\end{table}

\section{Experiments}

\begin{table*}
\centering
%   \label{tab:freq}

  \begin{tabular}{clccccc}
    \toprule
    \multicolumn{2}{c}{\multirow{3}{*}{Methods}} & \multirow{3}{*}{Features}
    ~ & \multicolumn{2}{c}{DrugBank} & \multicolumn{2}{c}{DGIdb}\\
    \cline{4-7}
    \multicolumn{2}{c}{~} & ~ & Validation & Ind. Test & Validation & Ind. Test\\
    \multicolumn{2}{c}{~} & ~ & ACC & ACC & ACC & ACC\\
    \midrule
    \multirow{3}{*}{MF-based}
    ~ & MC & no & - & 0.518 $\pm$ 0.013 & - & 0.559 $\pm$ 0.009 \\
    ~ & GRALS & yes & - & 0.532 $\pm$ 0.021 & - & 0.578 $\pm$ 0.016 \\
    ~ & F-EAE & no & - & 0.566 $\pm$ 0.004 & - & 0.623 $\pm$ 0.003 \\
    \midrule
    \multirow{4}{*}{GNN-based}
    ~ & GC-MC & yes & - & 0.586 $\pm$ 0.008 & - & 0.601 $\pm$ 0.005 \\
    ~ & sRGCNN & yes & - & 0.602 $\pm$ 0.010 & - & 0.689 $\pm$ 0.007 \\
    ~ & PinSage & yes & - & 0.629 $\pm$ 0.004 & - & 0.713 $\pm$ 0.005 \\
    ~ & IGMC & no & - & 0.634 $\pm$ 0.003 & - & 0.803 $\pm$ 0.006 \\
    \midrule
    % \cline{2-6}
    % \multirow{3}{*}{Aggregators}
    \multirow{7}{*}{Proposed}
    ~ & CoSMIG-w/GCN & no  & 0.562 $\pm$ 0.004 & 0.581 $\pm$ 0.004 & 0.778 $\pm$ 0.023 & 0.803 $\pm$ 0.009\\
    ~ & CoSMIG-w/GraphSAGE & no  & 0.584 $\pm$ 0.003 & 0.602 $\pm$ 0.008 & 0.807 $\pm$ 0.014 & 0.814 $\pm$ 0.010\\
    ~ & CoSMIG-w/RGCN & no  & 0.614 $\pm$ 0.004 & 0.637 $\pm$ 0.005 & 0.821 $\pm$ 0.013 & 0.832 $\pm$ 0.002\\
    % \midrule
    \cline{2-7}
    % \multirow{3}{*}{Pooling Layers}
    ~ & CoSMIG-w/AvgPooling & no  & 0.619 $\pm$ 0.003 & 0.643 $\pm$ 0.006 & 0.822 $\pm$ 0.006 & 0.835 $\pm$ 0.003\\
    ~ & CoSMIG-w/SumPooling & no  & 0.625 $\pm$ 0.004 & 0.655 $\pm$ 0.003 & 0.824 $\pm$ 0.007 & 0.839 $\pm$ 0.004\\
    ~ & CoSMIG-w/SortPooling & no  & 0.639 $\pm$ 0.002 & 0.667 $\pm$ 0.004 & 0.833 $\pm$ 0.003 & 0.841 $\pm$ 0.005\\
    % \midrule
    \cline{2-7}
    ~ & CoSMIG & no  & \textbf{0.658 $\pm$ 0.008}& \textbf{0.678 $\pm$ 0.003} & \textbf{0.840 $\pm$ 0.011}& \textbf{0.852 $\pm$ 0.012}\\
  \bottomrule
\end{tabular}
 \caption{The comparison of different methods by the overall accuracy on the DrugBank and DGIdb datasets in transductive scenario.}
\vspace{-1.0em}
\end{table*}

\subsection{Experimental Setup.}

\paragraph{Datasets.} To evaluate the effectiveness of CoSMIG, we compiled the multi-relational  datasets from DGIdb\cite{cotto2018dgidb} and DrugBank \cite{wishart2018drugbank}, respectively (Table 1). (1) DrugBank: This is the pharmaco-transcriptomics dataset obtained from DrugBank. Wherein, the up/down-regulation of genes due to the metabolism of pharmaceutical compounds represents the interactions between drugs and genes. Therefore, this dataset has two types of interactions, increased and decreased.  (2) DGIdb: This dataset is adopted from DGIdb, containing over 1664 genes and 1185 drugs involved in over 11,366 drug-gene interactions and 14 types of relationship. To ensure the quality of the dataset, we dropped duplicates and retained the drugs and genes with at least five interactions on the two datasets. %We summarized the statistics of datasets in .%We conduct experiments and split the dataset in two settings: inductive and transductive.

% \noindent\textbf{Data Splits.} We conduct experiments on the two datasets and split the dataset in two settings: inductive and transductive. For each transductive dataset, we randomly select 80\% of historical interactions of each drug to constitute the training set and treat the remaining as the test set. From the training set, we randomly select 10\% of interactions as validation set to tune hyper-parameters. Moreover, we randomly select 80\% of drugs and its interactions to constitute the training set to form the inductive dataset. As for the remaining drugs, we randomly select 20\% of interactions for each drug to constitute the initial subgraph for inductive inference and evaluate the remaining interaction for comparison. 

\noindent\textbf{Experimental Settings.} We implemented CoSMIG using pytorch geometric \cite{fey2019fast}, which is available at \url{https://github.com/biomed-AI/CoSMIG}. We tuned model hyperparameters based on cross validation results on DrugBank and used them across all datasets. The hop number h was set to 3. The depth of model was set to 4. For each subgraph, we randomly dropped out its adjacency matrix entries with a probability of 0.1 during the training. The training process lasted 80 epochs on a Nvidia GeForce RTX 3090 GPU. The impacts of several hyperparameters of our method such as embedding size and learning rate have been shown in Appendix Figure S1. As a multi-relational link prediction problem, we calculate the accuracy score (ACC) between the actual drug-gene interactions and the predictions to evaluate our method. We trained the model for five times and averaged the results to obtain the final performance.

\begin{table}
  \label{tab:freq}
  \resizebox{\linewidth}{18mm}{
  \begin{tabular}{cccccc}
    \hline
    \multicolumn{2}{c}{\multirow{2}{*}{Methods}} & \multicolumn{2}{c}{DrugBank} & \multicolumn{2}{c}{DGIdb}\\
    \cline{3-6}
    \multicolumn{2}{c}{~} & ACC & $\Delta$ & ACC & $\Delta$\\
    \hline
    \multirow{2}{*}{MF-based}
    ~ & IMC & 0.441 & 14.8\%  & 0.424 & 24.2\%\\
    ~ & F-EAE & 0.474 & 16.3\% & 0.532 & 14.6\%\\
    \hline
    \multirow{3}{*}{GNN-based}
    ~ & GC-MC & 0.513 & 12.4\% & 0.553 & 7.99\%\\
    ~ & PinSage & 0.567 & 9.86\% & 0.654 & 8.27\%\\
    ~ & IGMC & 0.612 & 3.47\% & 0.778 & 3.11\%\\
    \hline
    Proposed & CoSMIG & \textbf{0.672} & \textbf{0.88\%} & \textbf{0.842} & \textbf{1.17\%}\\
  \hline
\end{tabular}}
 \caption{Performance on the inductive scenario. $\Delta$ represents the decline rate between the transductive scenario and the inductive scenario of each model.}
\vspace{-1.0em}
\end{table}

% \subsection{Performance Comparison and Analysis}

% \begin{table}
%   \label{tab:commands}
%   \begin{tabular}{cccccc}
%     \hline
%     \multicolumn{2}{c}{Method}  &Features &DGIdb &DrugBank\\ \hline
%     \multirow{3}{*}{MF-based} & MC & no & 0.559 & 0.518\\ 
%     & GRALS & yes & 0.578 & 0.532\\ 
%     & F-EAE  & no & 0.623 & 0.566\\ 
%     \hline
%     \multirow{4}{*}{GNN-based} & GC-MC  & yes & 0.601& 0.586\\
%     & sRGCNN & yes & 0.689 & 0.602\\
%     & PinSage & yes & 0.713 & 0.629\\
%     & IGMC & no & 0.803 & 0.634\\ 
%     \hline
%     Proposed & CoSMIG & no & \textbf{0.852} & \textbf{0.678} \\ 
%     \hline
%   \end{tabular}
%   \caption{Accuracy test scores on transductive dataset}
% \end{table}

\subsection{Performance Comparison.}

\paragraph{Performance in transductive scenario.} The compared methods included three MF-based (MC, GRALS, F-EAE) and five GNN-based (GC-MC, sRGCNN, PinSage, IGMC) methods. As shown in Table 2, CoSMIG achieved the highest accuracies on both the DrugBank and DGIdb. The poor performance of MF-based methods (MC, GRALS, F-EAE) may be ascribed to the fact that the simple low rank (or sparse) matrix approximation is insufficient to capture the complex interaction types, whereas the performance of GC-MC and sRGCNN verify that the node-level of graph neural network can improve the representation learning and benefit the prediction of drug-gene interactions. IGMC, instead of learning transductive node-level features, learns local graph patterns related to the interactions inductively based on relational graph neural network (R-GCN), showing highly competitive performance compared to the other baselines. Nevertheless, strengthening the role of interaction embeddings in the subgraph modeling, CoSMIG improves over IGMC by 6.49\% and 5.75\% on the DrugBank and DGIdb, respectively. Furthermore, CoSMIG continues to outperform other baseline methods when evaluating on each interaction type (Fig. 3A).

% \subsection{CoSMIG is superior to other methods in inductive scenario.}

\paragraph{Performance in inductive scenario.} Note that GRALS and sRGCNN could not be applied to the inductive scenario due to their model settings. As shown in Table 3, CoSMIG consistently outperformed other baselines on the inductive datasets. The MF-based methods still achieved the poorest performances in inductive scenario and there is a huge performance gap between the transductive and inductive scenario.  Moreover, IGC-MC and PinSage also show a significant decrease (the rate of decline ranging from 7.99\% to 12.4\%) in the inductive scenario, likely due to their only node-level message passing mechanism. Compared to the node-level GNN methods, IGMC decreases slightly because their extracted enclosing subgraph contains rich graph pattern information without any global information in the drug-gene graph. Nevertheless, CoSMIG achieves consistent outperforms IGMC by 8.93\% and 7.60\% on the DrugBank and DGIdb, respectively. And the decline rate of CoSMIG is significantly lower than other methods. The remarkable improvement can be attributed to its RWR subgraph extraction methods and communicative message passing mechanism.

\begin{figure}[h]
  \centering
  \includegraphics[width=1.0\linewidth]{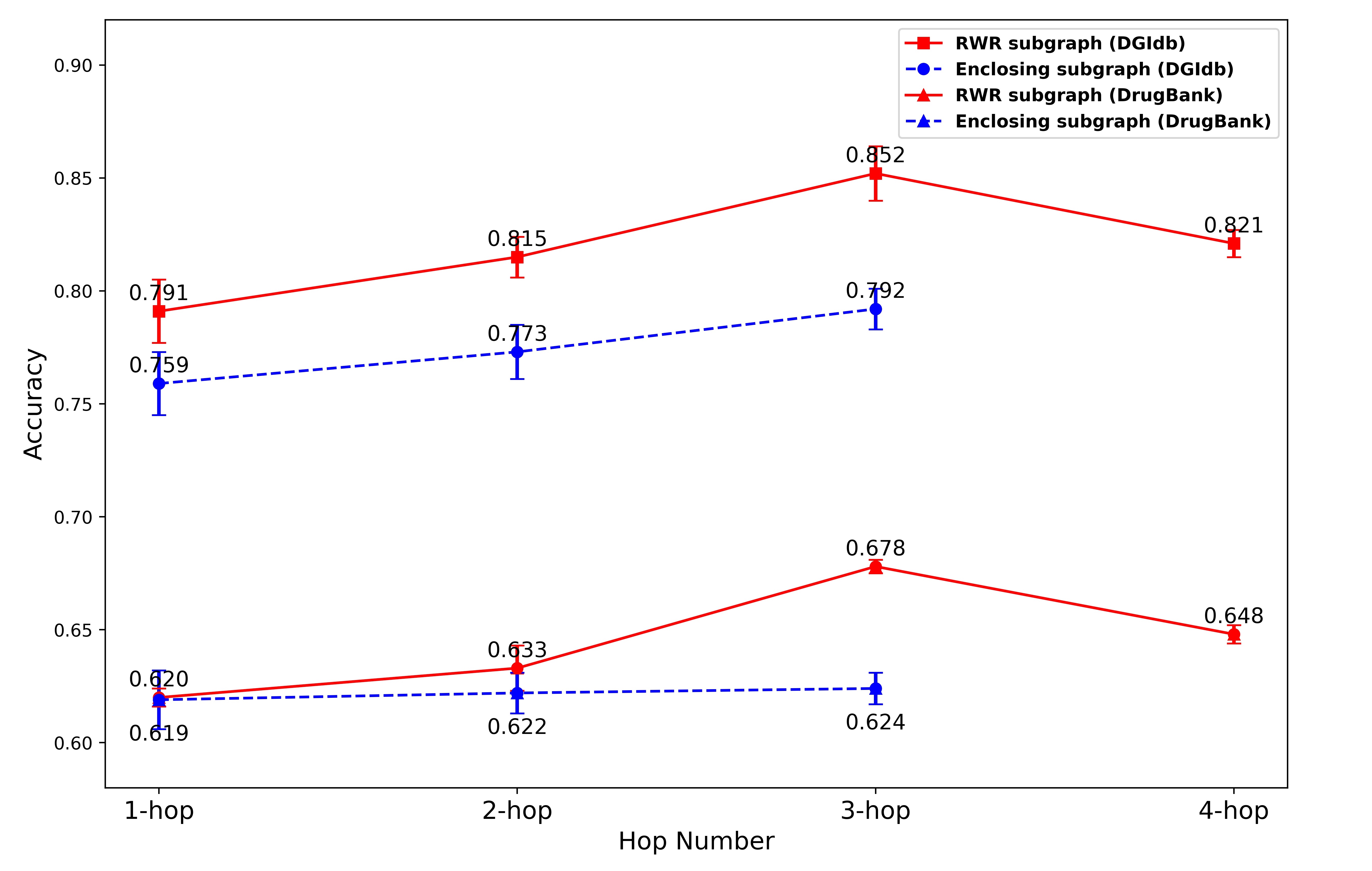}
  \caption{Evaluating the effect of Subgraph Extraction on DrugBank and DGIdb.}
\end{figure}

\subsection{Ablation Study.}

We conducted ablation studies on two benchmark datasets to investigate factors that influence our performance. As shown in Table 2, when using different types of GNN aggregators, all variants were inferior to all COSMIG. The best baseline,  CoSMIG-RGCN, obtained accuracy score of 0.821 and 0.832 for the validation and test set on DGIdb, which are 2.26\% and 2.35\% lower than our proposed model. For three pooling layer variants, CoSMIG-w/SortPooling performs the best, while its accuracy is 0.83\% and 1.29\% lower than CoSMIG on the validation set and test-set of DGIdb, respectively. The consistent results between the validation and independent test sets also indicated the robustness of the CoSMIG model.

To indicate the influence of subgraph extraction, we also compared our RWR subgraph extraction with the enclosing subgraph extraction method proposed in IGMC. And we varied the hop number of neighbors to investigate whether CoSMIG can benefit from multiple hop neighbors. The results are shown in Figure 2. We observe that the performance of RWR subgraph extraction method consistently outperforms the enclosing subgraph method both on the DrugBank and DGIdb datasets. And as the hop number increases, the Accuracy of CoSMIG increases rapidly due to the more informative neighbors and interactions in the subgraph, and the model reaches its best performance at the hop number of 3. Subsequently, the performance slowly descends as the hop number continues to increase, indicating that an overlarge hop number will bring redundant and harmful information.

\begin{figure}[h]
      \centering
	   \includegraphics[width=1.0\linewidth]{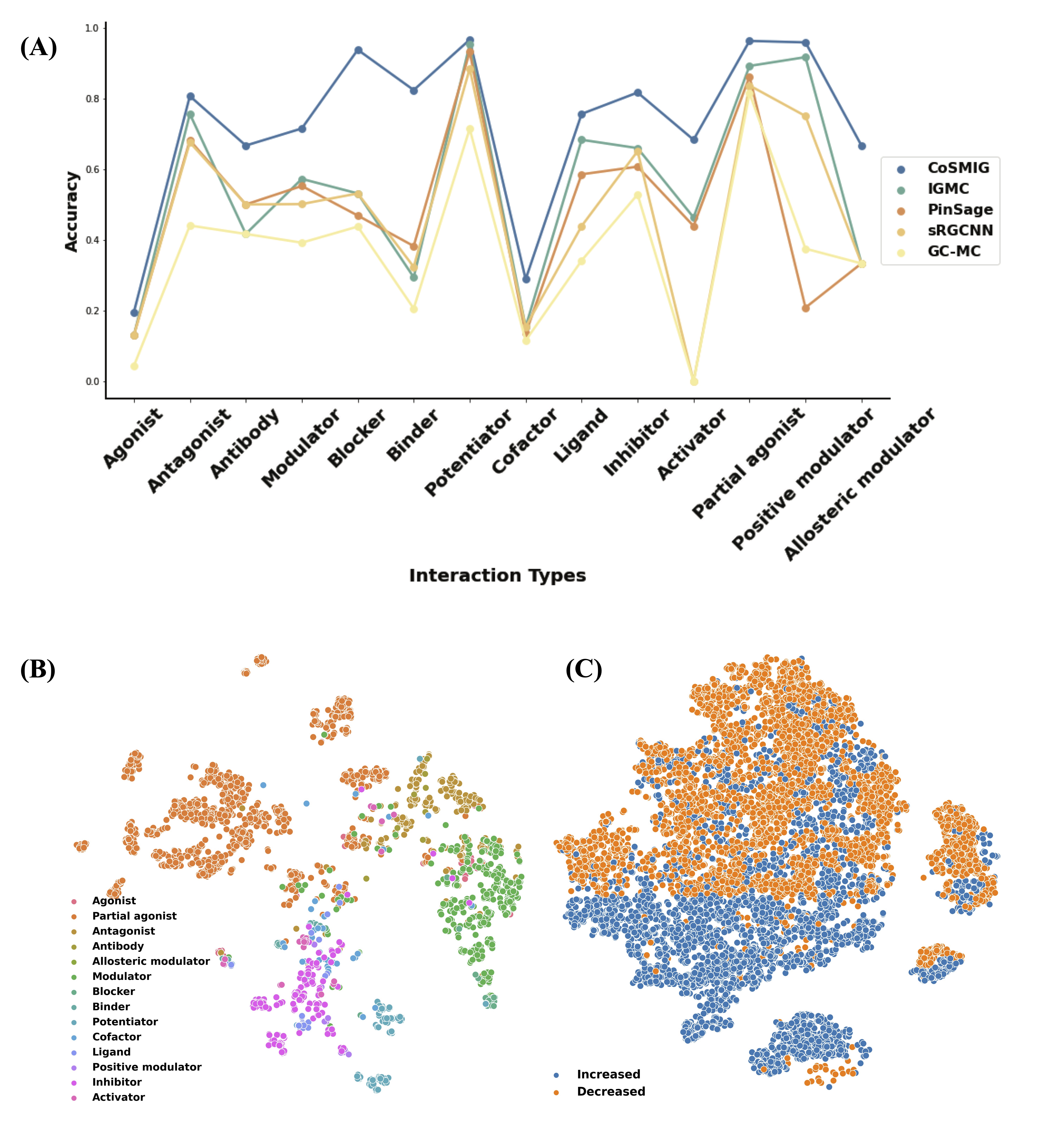}
	   \caption{(A) The accuracy of each interaction type for DGIdb, and the pairs projected by t-SNE for (B) DGIdb and (C) DrugBank.}
\end{figure}
% \begin{figure}[h]
%   \centering
%   \includegraphics[width=1.0\linewidth]{Figure 5.jpg}
%   \caption{Real Example of novel drug-gene interactions predicted by CoSMIG.}
% \end{figure}

\subsection{Experimental Analysis.}

\paragraph{Visualization Analysis.} We extracted the learned representations of drug-gene interaction pairs from the trained CoSMIG model, and project them into 2D space using tSNE as shown in Figure 3B-C. Obviously, CoSMIG could clearly distinguish each interaction type of DGIs, indicating that CoSMIG allows accurate representations for DGIs predictions.

\paragraph{Validation on the LINCS L1000 database.} We further performed large-scale computation of up/down regulating genes and evaluated the predictions through the LINCS L1000 database. As detailed in Appendix Section 2, these results highlight the superiority of CoSMIG and our predicted increase/decrease of gene expression caused by drugs are highly consistent with the experimental values.

\paragraph{Discovering novel drug-gene interactions.}  We also demonstrated the new potential interactions that are not in the current datasets but have been reported in the literature as detailed in Appendix Section 3.

\section{Conclusion}

In this work, we firstly compiled two new challenging benchmarks for multi-relational drug-gene interactions prediction instead of focusing solely on the binding interaction. Furthermore, our model is naturally inductive based on the communicative subgraph neural network without leveraging any domain knowledge and global information specific to the drug-gene graph. Therefore, our model could be applied to unseen drugs/genes without retraining. Finally, we conducted extensive experiments to demonstrate the effectiveness of our model and its interpretability in understanding the multiple relational information. To demonstrate the value of our model, we evaluated the interactions types of increased and decreased between drugs and genes with the gene expression profile from LINCS L1000 database. Moreover, we demonstrated the new potential interactions that are not in the current datasets but have been reported in the literature.

% Note that our model still requires an unseen drugs/genes subgraph (i.e., the gene and drug should at least have some interactions with neighbors so that the subgraph is not empty). This scenario is more common in practice. For example, when performing drug repositioning, the drug candidate usually has a few known interactions with known genes based on experimental results. In this case, CoSMIG can be of great value by still making predictions based purely on known interactions with genes. 

% A future direction of our work is to integrate the domain knowledge with the subgraph representation in our framework. While we used only drug-gene bipartite graph in this work, we highlight that CoSMIG is a scalable framework in that more additional networks such as drug–drug interactions and protein-protein interactions can be easily incorporated into the current framework.

\section*{Acknowledgments}

This study has been supported by the National Key R\&D Program of China [2020YFB0204803], National Natural Science Foundation of China [61772566, 62041209], Guangdong Key Field R\&D Plan [2019B020228001, 2018B010109006], Introducing Innovative and Entrepreneurial Teams [2016ZT06D211], and Guangzhou S\&T Research Plan [202007030010].

% \section{\LaTeX{} and Word Style Files}\label{stylefiles}

% The \LaTeX{} and Word style files are available on the IJCAI--22
% website, \url{https://ijcai-22.org/}.
% These style files implement the formatting instructions in this
% document.

% The \LaTeX{} files are {\tt ijcai22.sty} and {\tt ijcai22.tex}, and
% the Bib\TeX{} files are {\tt named.bst} and {\tt ijcai22.bib}. The
% \LaTeX{} style file is for version 2e of \LaTeX{}, and the Bib\TeX{}
% style file is for version 0.99c of Bib\TeX{} ({\em not} version
% 0.98i). The {\tt ijcai22.sty} style differs from the {\tt
% ijcai21.sty} file used for IJCAI--21.

% The Microsoft Word style file consists of a single file, {\tt
% ijcai22.docx}. This template differs from the one used for
% IJCAI--21.

% These Microsoft Word and \LaTeX{} files contain the source of the
% present document and may serve as a formatting sample.

% Further information on using these styles for the preparation of
% papers for IJCAI--22 can be obtained by contacting {\tt
% proceedings@ijcai.org}.

\bibliographystyle{named}
\bibliography{ijcai22}

\end{document}

\maketitle
\renewcommand{\thefootnote}{\fnsymbol{footnote}}
\footnotetext[1]{Corresponding authors.}
\footnotetext[2]{Work done during an internship at Galixir.}

\section{Experiments}

\subsection{Data Splits}

We conduct experiments on the two datasets (DrugBank and DGIdb) and split the dataset in two settings: inductive and transductive. For each transductive dataset, we randomly select 80\% of historical interactions of each drug to constitute the training set and treat the remaining as the test set. From the training set, we randomly select 10\% of interactions as validation set to tune hyper-parameters. Moreover, we randomly select 80\% of drugs and its interactions to constitute the training set to form the inductive dataset. As for the remaining drugs, we randomly select 20\% of interactions for each drug to constitute the initial subgraph for inductive inference and evaluate the remaining interaction for comparison.

% \begin{figure*}[h]
%   \centering
%   \includegraphics[width=0.9\linewidth]{figS1.jpg}
%   \caption{Parameter Analysis. (A) Embedding Size and (B) Learning rate}
% \end{figure*}

% \begin{figure}
%   \centering
%   \includegraphics[width=0.9\linewidth]{FigS1.eps}
%   \caption{Real Example of novel drug-gene interactions predicted by CoSMIG.}
% \end{figure}

\subsection{Hyper-parameters Search}
In this section, we will analyze the impacts of several hyper-parameters of our method such as embedding size and learning rate. In particular, we vary the embedding size of CoSMIG in \{16, 32, 64, 128\} to investigate the efficiency of the embedding propagation layer. And the learning rate is also searched in the range of \{0.001, 0.0001, 0.00001\}.

From Figure S1(A), we could see that different embedding sizes would result in different final scores but a small volatility rate. It hence illustrates the stability and effectiveness of our method, that the test results would not fluctuate too much if the embedding size is in an appropriate range. Furthermore, in our comparison of different learning rate (Fig. S1(B)), we could find that if the learning rate is too small, the final results would be stuck in a poor local minimum. It again shows that our method could achieve stable final results.

\begin{figure}[h]
	\subfigure[Embedding Size]{
	\begin{minipage}{4cm}
	\includegraphics[scale=0.25]{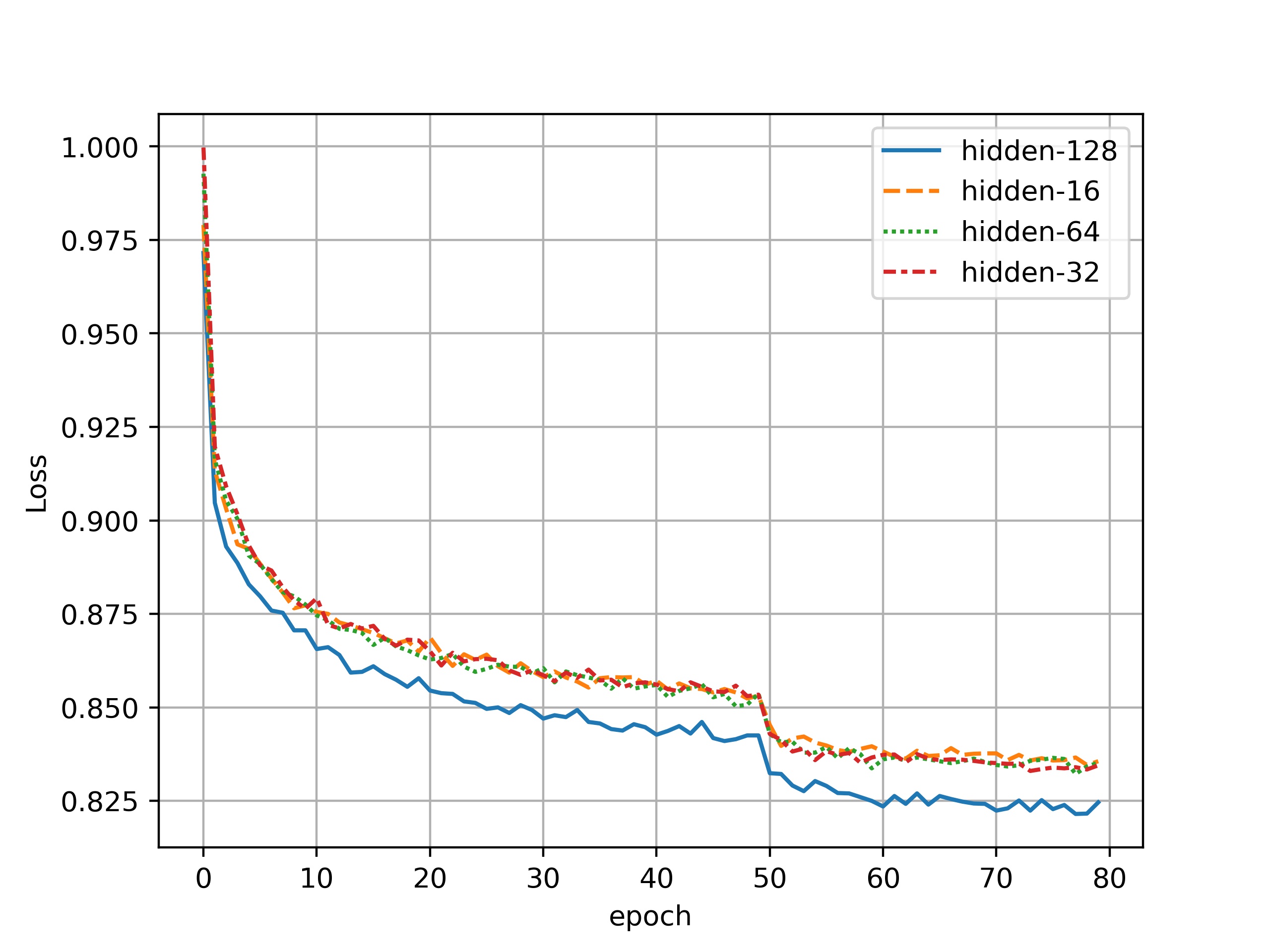}        
	\end{minipage}
	}
	\subfigure[Learning rate]{          
	\begin{minipage}{4cm}
	\includegraphics[scale=0.25]{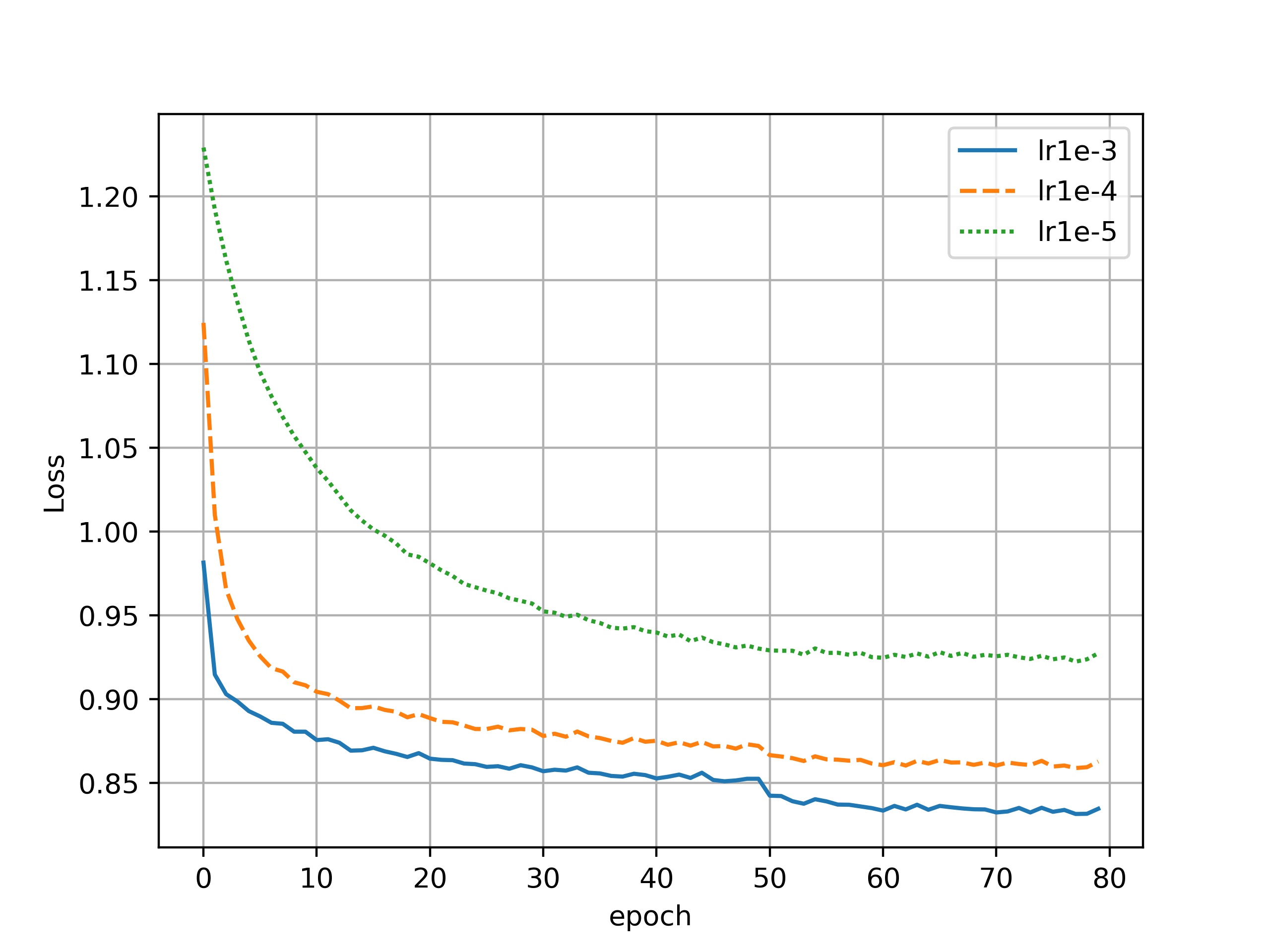}        
	\end{minipage}
	}
	\caption{Parameter Analysis} 
\end{figure}

\begin{figure*}[h]
  \centering
  \includegraphics[width=1.0\linewidth]{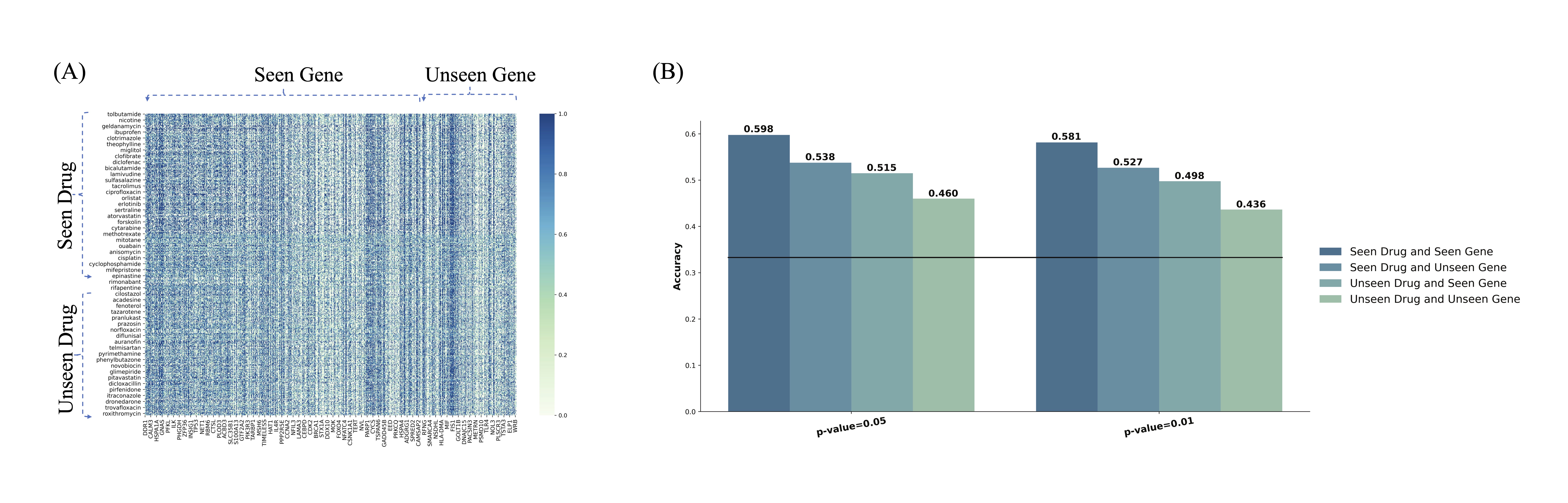}
  \caption{(A) The P-value calculated from LINCS L1000 database, that used for ground truth. (B) The performance of CoSMIG on the external test-set at different p-value thresholds. DGIs from LINCS were divided into four groups based on their presence in the training set.}
\end{figure*}

\section{Validation on LINCS L1000 Database}

By measuring the gene expression changes for \~1000 genes perturbed by \~30000 compounds, the LINCS database provided new insight into the influence of drugs on genes, such as the up/down-regulation of genes. Therefore, we applied t-test on the gene expression data to determine whether the expression changes of a gene perturbed by a drug is significantly greater/less than 0, corresponding to the interaction type of increased and decreased. Note that we selected 978 landmark genes and 403 drugs from LINCS database to constitute the external test set, including 932 genes and 203 drugs seen in the training set and 46 genes and 201 drugs unseen.

We further performed large-scale computation of up/down regulating genes and evaluated the predictions through the LINCS L1000. As shown in Figure S2(A), all drugs and genes have the same pattern, suggesting that the unreliable gene expression profiles in LINCS L1000 have not been included. The horizontal line is defined as the randomness of the model predictions. Figure S2(B) showed the accuracy of CoSMIG at different p-value thresholds. At the threshold of 0.05, CoSMIG obtained the accuracy of 0.598, 0.538, 0.515, 0.460 in each group that significantly outperformed the random baseline. The remarkable improvement over random baseline again indicates that CoSMIG has learned the complex correlations of drugs and genes, which have been evaluated by the gene expression profiles from LINCS L1000. Furthermore, the accuracy of the unseen drugs and the unseen genes also improves over random baseline by 39.39\%, demonstrating that our model is superior on the inductive scenario. These results highlight the superiority of CoSMIG in predicting the interaction types between drugs and genes in transductive and inductive scenarios.

\begin{table*}
    \centering
    \begin{tabular}{cccc}
    \hline
    Drug  & Gene & Predicted interaction & Literature \\
    \hline
    Metformin & CNR2  & partial agonist &   \cite{dailey2008beyond,rotella2008time}   \\
    Metformin & CHRNB4  & antagonist &       \\
    Metformin & ITPR2  & antagonist &   \cite{tubbs2014mitochondria}   \\
    Metformin & P4HA1  & partial agonist &      \\
    Metformin & KDM5D  & antagonist &   \cite{sharma2020kdm6a}   \\
    Metformin & P3H3  & blocker &      \\
    Metformin & HTR1A  & binder &      \\
    Metformin & PGR  & blocker &  \cite{giles2012obesity,wu20115}    \\
    Metformin & CLIC1  & modulator &    \cite{liu2017chloride}  \\
    Metformin & PPAPG  & potentiator &   \cite{natali2006effects,tiikkainen2004effects}   \\
    \hline
    Atropine & MAPT  & antagonist &   \cite{hellstrom2000increased}   \\
    Apomorphine & MAPT  & blocker &       \\
    Dopamine & MAPT  & binder &  \cite{klein2005tau,klein2008tau}    \\
    Decamethonium & MAPT  & partial agonist &      \\
    Cyproheptadine & MAPT  & partial agonist &      \\
    Etoposide & MAPT  & blocker &   \cite{suuronen2000protective,furgerson2012model}   \\
    Cabergoline & MAPT  & modulator &      \\
    Tetrahydrocannabinol & MAPT  & cofactor &      \\
    Cyproterone acetate & MAPT  & activator &   \cite{butler2001androgen}  \\
    Methylphenobarbital & MAPT & antagonist & \\
    Metergoline & MAPT  & partial agonist &   \\
    \hline
    \end{tabular}
    \caption{Summary of novel DGIs predictions.}
    \label{tab:plain}
\end{table*}

\begin{figure}[h]
  \centering
  \includegraphics[width=1.0\linewidth]{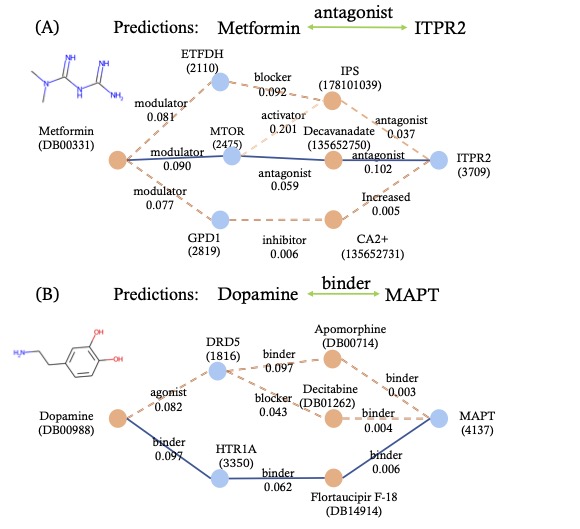}
  \caption{Real Example of novel drug-gene interactions predicted by CoSMIG.}
\end{figure}

\section{Discovering noval drug-gene interactions}
To further validate the prediction ability of CoSMIG, we conducted case studies for the diabetes medicine: Metformin, and the neurodegenerative diseases related gene: MAPT. Supplementary Table 1 shows the novel DGI predictions, with the canonical name and identifier of the drug, predicted interaction, gene name, and literature reference supporting interpretation. As shown in Figure S3, CoSMIG captures the high-order connectivity and complex relation information, which play a key role to infer DGIs.

For the diabetes medicine Metformin, among the top 10 predicted DGIs ranked according to their prediction scores, 6 DGIs (60\% success rate) were validated by literature evidence in which there is literature evidence indicating the possible interaction between the drug and the gene. For example, Metformin is predicted by CoSMIG to be the partial agonist of gene CNR2. A previous clinical study reported that there is an association of the cannabinoid receptor (CNR) gene with the obese patients with diabetes mellitus type 2 after Metformin treatment \cite{dailey2008beyond,de2014role}. More importantly, the supporting studies also have shown that the cannabinoid receptor (CNR) antagonists, may have favorable effects on type-2 diabetes \cite{de2014role}. Furthermore, another prediction between Metformin and gene ITPR2 was also supported by a previous study indicating that CoSMIG is capable of discovering the novel DGIs \cite{tubbs2014mitochondria}. 

For the neurodegenerative diseases related gene MAPT, we also focused on the top-5 CoSMIG-predicted candidates in Supplementary Table 1. We found that 4 of 10 drugs (40\% success rate) were validated by previous studies from literatures. Herein, Dopamine, a compound of the catecholamine and phenethylamine families playing important roles in the human brain, was predicted by CoSMIG to be associated with MAPT. Such a prediction can be supported by a previous study indicating that Dopamine neuron function is affected by tau gene transfer \cite{klein2005tau,klein2008tau}.

\section{Discussion}
In this work, we firstly compiled two new challenging benchmarks for multi-relational drug-gene interactions prediction and proposed a novel inductive subgraph representation learning framework for accurate DGI predictions. 

Note that our model still requires an unseen drugs/genes subgraph (i.e., the gene and drug should at least have some interactions with neighbors so that the subgraph is not empty). This scenario is more common in practice. For example, when performing drug repositioning, the drug candidate usually has a few known interactions with known genes based on experimental results. In this case, CoSMIG can be of great value by still making predictions based purely on known interactions with genes. 

A future direction of our work is to integrate the domain knowledge with the subgraph representation in our framework. While we used only drug-gene bipartite graph in this work, we highlight that CoSMIG is a scalable framework in that more additional networks such as drug–drug interactions and protein-protein interactions can be easily incorporated into the current framework.

\bibliographystyle{named}
\bibliography{ijcai22}